\documentclass{bvm} 

\usepackage{todonotes}

\begin{document}

\newcommand{\bvmyear}{2023}

\selectlanguage{english} 

\title{Combining Image- and Geometric-based Deep Learning for Shape Regression}

\subtitle{A Comparison to Pixel-level Methods for Segmentation in Chest X-Ray}

\titlerunning{Learning-based Shape Regression}

\author{
	Ron \lname{Keuth}, 
	Mattias~P. \lname{Heinrich} 
}

\authorrunning{Keuth \& Heinrich}

\institute{Institute of Medical Informatics, University of Lübeck}

\email{r.keuth@uni-luebeck.de}

\maketitle

\begin{abstract}
When solving a segmentation task, shaped-base methods can be beneficial compared to pixelwise classification due to geometric understanding of the target object as shape, preventing the generation of anatomical implausible predictions in particular for corrupted data. In this work, we propose a novel hybrid method that combines a lightweight CNN backbone with a geometric neural network (Point Transformer) for shape regression. Using the same CNN encoder, the Point Transformer reaches segmentation quality on per with current state-of-the-art convolutional decoders ($4\pm1.9$ vs $3.9\pm2.9$ error in mm and $85\pm13$ vs $88\pm10$ Dice), but crucially, is more stable w.r.t image distortion, starting to outperform them at a corruption level of 30\%.
Furthermore, we include the nnU-Net as an upper baseline, which has $3.7\times$ more trainable parameters than our proposed method.
\end{abstract}

\section{Introduction}
Semantic segmentation is a fundamental task of medical image processing. It identifies an image object by classifying each pixel or aligning a shape model comprising anatomical landmarks from a known image to a new one. Recently, deep learning has set a new state-of-the-art in this field, especially the U-Net with its self-adapting nnU-Net framework \cite{isensee_nnu-net_2021} for the pixel-wise classification approach. However, tackling the shape alignment approach with deep learning is also an active field of research, motivated by the geometric understanding of the object itself, which the pixel-wise classification lacks. This offers several advantages, like addressing limitations such as anatomical implausible predictions of convolutional decoders or opening the possibility of human interaction by propagating local refinements through the whole predicted shape.
In \cite{castrejon_annotating_2017}, ConvLSTM cells are used to predict contour points sequentially on the features of a CNN backbone, allowing refinement by the user at each step. Newer work combines such CNN backbones with graph neural networks. In PolyTransform \cite{liang_polytransform_2020}, the contour of a predicted segmentation mask is refined by a transformer block, and \cite{gaggion_improving_2023} combines a CNN encoder and a graph convolutional decoder with a new type of image-to-graph skip connections.
In this work, we propose a shape-based segmentation pipeline that also utilizes geometric deep learning, yielding promising results using a much smaller CNN encoder (preventing the risk of overfitting) compared to the nnU-Net (factor: 3.7). We formulate the shape-regression in three different approaches and furthermore reveal its higher robustness for corrupted input compared to the pixel-level baselines.

\section{Methods}

\begin{figure}[b]
	\centering
    \includegraphics[width=\figwidth]{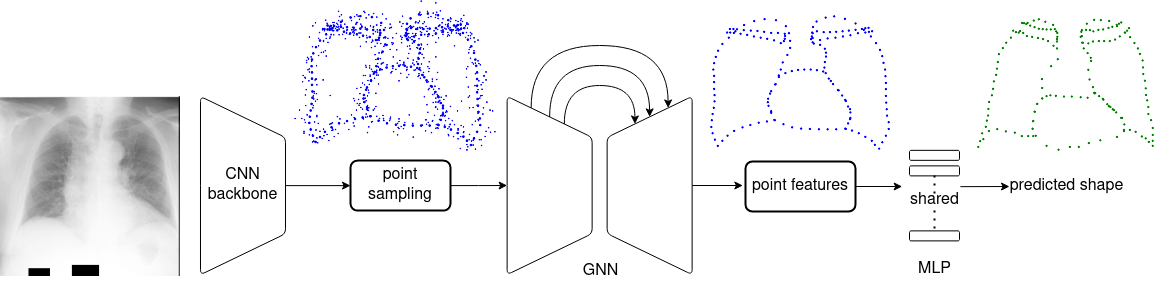}
	\caption{A schematic overview of our pipeline for shape regression: A pretrained CNN backbone extracts image features, which are sampled in a point cloud for the geometric neural network (GNN) using a random initial shape of the training data. A shared MLP head finally predicts the shape directly or via relative displacement for each landmark.}
	\label{fig:schema_pipeline}
\end{figure}

\subsection{Pipeline for shape regression} \label{sec:pipeline_introduction}
Our proposed pipeline consists of multiple parts (Fig. \ref{fig:schema_pipeline}): a Convolutional Neural Network (CNN) backbone extracts features from the input image, followed by a Graph Neural Network (GNN), and finally, a Multilayer Perceptron (MLP) that maps the generated point features to the desired solution space. To define a 2D point cloud for shape regression, we start from an initial shape (randomly picked from another training sample) and add four additional points for each landmark's local neighbourhood, which were obtained by sampling its offsets from a normal distribution with a small standard derivation. We then sample (bilinear interpolation) image features from the respective positions.

As a CNN-backbone, we adopt an ImageNet-pretrained ResNet18 \cite{He2016} by cropping it to the first 12 layers and replacing the last two convolutions with dilated convolution to obtain a higher resolution feature map of $32\times32$ pixels with $64$ channels.

We explore two GNN architectures in our pipeline. As a baseline, we chose the well-established PointNet \cite{qi_pointnet_2016} and the Point Transformer \cite{zhao_point_2021}, which applies the transformer's self-attention mechanism to the local neighbourhood for each point in the point cloud. This enables us to compare the PointNet`s performance to the new state-of-the-art.
In our pipeline, the GNN takes the $L\cdot5=830$ landmarks, including their additional four sampling points, and projects each in a $K=128$ feature space.
We then select only those point features that correspond to the $L$ landmarks of the initial shape and feed the $L\times K$ feature matrix to the MLP head, which generates the final prediction. This MLP head is shared for all $L$ points and consists of three layers doubling $K$ and projecting them to the desired dimensionality $M$, where $M$ differs depending on the employed type of shape regression.

All our models are trained with the same hyperparameters: We use the Adam optimiser (lr=$1e-3$) and reduce its learning with factor $0.1$, if the validation loss has not decreased in the last 30 epochs. We train our models in a supervised manner, reducing the $L2$ distance (first term of Eq. \ref{eq:displacement}) between the predicted and ground truth shape for 500 epochs. As evaluation metrics, we chose the Dice Similarity Coefficient (DSC) and the Average Surface Distance (ASD) in mm and average over five random initial shapes.

\subsubsection{Formulation of shape regression}\label{sec:shape_approaches}
In the following, we describe our three different formulations of shape regression.

With the first method, $M=2$ holds the \emph{relative displacements} in $x$ and $y$ direction for the current landmark. We limit the possible value space to a local neighbourhood of $45\times45$ pixels (in the image space), and we require a smooth displacement field, which is added to the $L2$ distance loss as regularisation:
\begin{equation}\label{eq:displacement}
    \mathcal{L}_\text{disp} = (1-\lambda_\mathcal{R}) ||S^*- S_\text{init} + U||_2^2 + \lambda_\mathcal{R} ||\nabla U||^2\text{,}
\end{equation}
with $S^*,S_\text{init}\in\mathbb{R}^{L\times2}$ describing the ground truth and randomly picked initial shape with its relative displacement field $U\in\mathbb{R}^{L\times2}$. We use finite differences to calculate $\nabla U$. $\lambda_\mathcal{R}\in[0,1]$ weights the impact of the regularisation term. In our experiments, we obtain the best results with $\lambda_\mathcal{R}=0.2$.

Secondly, we formulate the shape regression as a \emph{heatmap regression} \cite{leibe_human_2016}. For this, we define the range of potential displacements as the same local neighbourhood as in the previous approach. We cover this neighbourhood with an $11\times11=121=M$ regular grid $G$ and let the model predict the likelihood of each grid point being the new position of the landmark. To obtain the likelihood, we normalize the logits with a $\texttt{softmax}$ and calculate the relative displacement via the weighted sum over all grid points' coordinates in $G$.

As a last approach, we formulate the problem as a \emph{direct shape regression}. Therefore, the final MLP head predicts the likelihood ($M=160$) for each of the 160 training shapes given the image and initial shape. We use these likelihoods as weights in a linear combination of all training shapes to generate the predicted shape.

\subsection{Pixel-level baselines}
For a comparison to segmentation using a pixel-wise classification, we chose two well-established deep learning architectures. As upper baseline, we use the nnU-Net framework \cite{isensee_nnu-net_2021} to train two U-Nets on our training split (with the \texttt{all} fold option) to avoid overlapping of lungs and clavicles. 
However, to make a direct comparison using the same feature input, we additionally train an LR-ASPP head \cite{howard_searching_2019} on the same pretrained CNN backbone, but omit the skip connection from the intermediate encoder layer to make the comparison fair. We train it using the same hyperparameters as our pipeline but replace the $L2$ loss with a weighted binary cross entropy loss.

\section{Results}
\begin{table}[t]
	\caption{The average surface distance (ASD) in mm and the Dice Similarity Score (DSC) in percent for the 87 test images (using five initial shapes). Point Transformer and PointNet are used as GNN in our pipeline (Fig. \ref{fig:schema_pipeline}) with the three regression approaches: displacement, heatmap, and shape (Sec. \ref{sec:shape_approaches}). The best result for shape-based and pixel-level approaches is highlighted in bold.}
	\label{tab:quantitave_results}
	\begin{tabular*}{\textwidth}{@{\extracolsep{\fill}}lcccccccc}
		\hline
         & & \multicolumn{3}{c}{Point Transformer} & PointNet & mean shape & \multicolumn{2}{c}{pixel-level baseline}\\
        structure               & metric    & disp & heatmap & shape & disp & & LR-ASPP & nnU-Net\\\hline
        \multirow[c]{2}{*}{lungs} & ASD     &3.5±1.5&3.6±1.7&6.7±3.1&3.8±1.7&11.4±5.1&2.9±2.1&1.4±0.6\\
        & DSC                               &94.5±2.7&94.2±2.9&89.2±5.5&93.9±3.0&82.1±7.4&96.2±0.9&98.1±0.8\\
        \multirow[c]{2}{*}{heart} & ASD     &5.3±2.3&5.3±2.7&9.0±3.7&5.8±3.0&13.2±7.0&4.5±3.3&3.0±1.2\\
        & DSC                               &91.8±3.9&91.7±4.3&84.8±6.7&91.1±5.0&77.2±12.0&93.6±2.2&95.2±2.1\\
        \multirow[c]{2}{*}{clavicles} & ASD &3.8±1.7&3.9±1.7&5.0±2.2&4.6±2.1&7.8±3.8&4.5±3.2&1.1±0.3\\
        & DSC                               &73.0±11.6&71.8±11.2&58.5±22.1&61.3±16.1&37.8±27.0&76.8±5.4&93.4±2.7\\\hline
        \multirow[c]{2}{*}{average} & ASD   &\textbf{4.0±1.9}&4.1±2.0&6.5±3.3&4.5±2.3&10.3±5.5&3.9±2.9&\textbf{1.6±1.0}\\
        & DSC                               &\textbf{85.3±12.7}&84.8±13.0&76.0±20.6&80.3±18.8&63.4±28.0&87.9±9.8&\textbf{95.6±2.9}\\
		\hline
	\end{tabular*}
\end{table}

\begin{figure}[b]
    \setlength{\figwidth}{.45\textwidth}
    \centering
    \begin{subfigure}[t]{\figwidth}
        \centering
		\includegraphics[width=.75\textwidth]{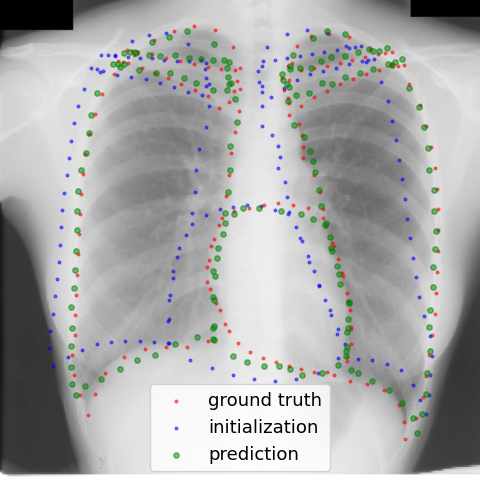}
		\caption{JSRT example}
		\label{fig:jsrt_example}
	\end{subfigure}
    \hfill
    \begin{subfigure}[t]{\figwidth}
        \centering
		\includegraphics[width=\textwidth]{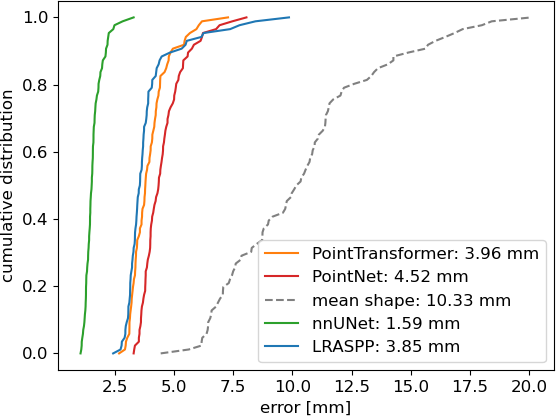}
		\caption{Distribution of the ASD}
		\label{fig:asd_distribution}
	\end{subfigure}
    \begin{subfigure}[t]{\figwidth}
        \centering
        \includegraphics[width=\textwidth]{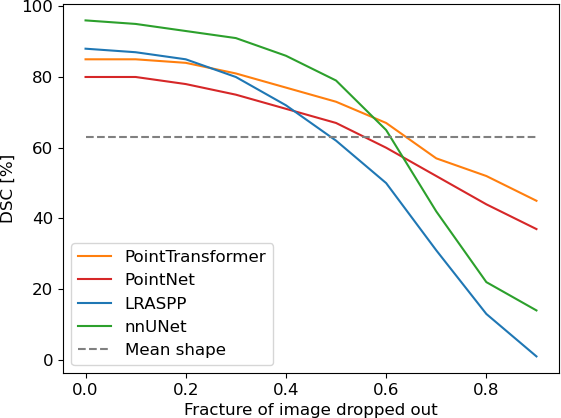}
        \caption{Ablation study: absolute performance}
		\label{fig:ablation_abs_perf}
    \end{subfigure}
    \hfill
    \begin{subfigure}[t]{\figwidth}
        \centering
        \includegraphics[width=\textwidth]{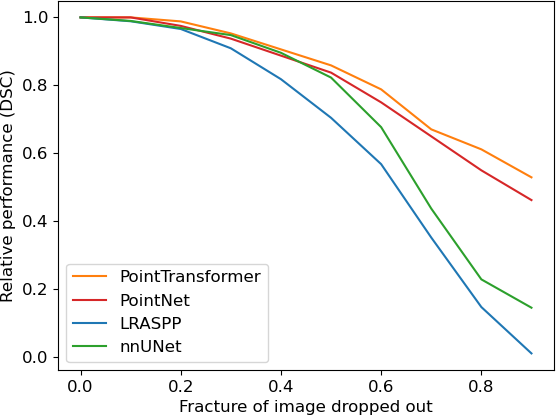}
        \caption{Ablation study: relative performance}
		\label{fig:ablation_rel_perf}
    \end{subfigure}
    \caption{Top left, a test image showing the initial, ground truth and predicted shape. Average surface distance (ASD) for the test split (top right) and results of our ablation study (bottom).}
    \label{fig:result_statistic_plots}
\end{figure}

\begin{SCfigure}

	\centering 
	
	\setlength{\figwidth}{0.3\textwidth}
	\begin{subfigure}{\figwidth}
		\includegraphics[width=\textwidth]{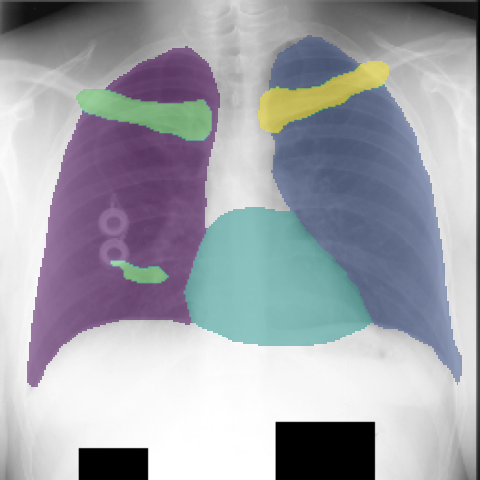} 
		\caption{LR-ASPP}
	\end{subfigure}
	\hfill	
	\begin{subfigure}{\figwidth}
		\includegraphics[width=\textwidth]{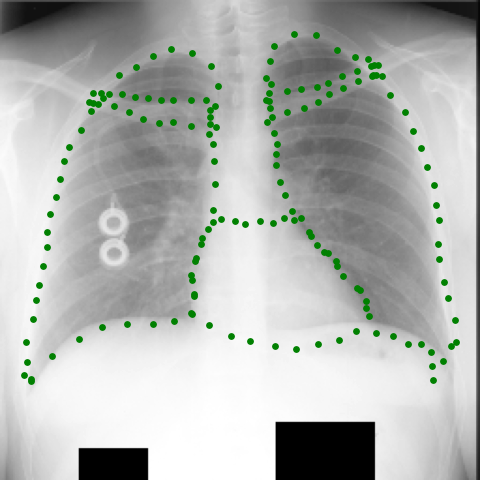}
		\caption{Point Transformer}
	\end{subfigure}
	\caption{Predictions of a test image showing an advantage of shape-based segmentation. The two ports in the left lung cause a domain shift and result in a false positive error by the pixelwise classification. In contrast, such anatomical implausibilities cannot be generated with a shape-based approach.}
	\label{fig:port_artefact}
\end{SCfigure}

We train and test our methods on the Japanese Society of Radiological Technology (JSRT) Dataset \cite{shiraishi_development_2000}. The JSRT consists of 247 chest X-rays with a resolution of $1024\times1024$ pixels and an isotropic pixel spacing of $0.175\ts\text{mm}$. Each image comes with human expert landmark annotations ($L=166$) for the four anatomical structures of the right lung (44), left lung (50), heart (26), right clavicle (23), and left clavicle (23) (Fig. \ref{fig:jsrt_example}). For our experiments, we downsample the dataset to $256\times256$ pixels and divide it into a custom split of 160 training and 87 test images. We use the mean shape of the training split as lower baseline in our experiments.

In general, the quantitative results in Tab. \ref{tab:quantitave_results} show that 
the pixel-level baselines outperform the graph-based models, with the nnU-Net as the overall best performance, with an ASD of $1.6\pm1\ts\text{mm}$ and a DSC of $96\pm3\ts\%$. Comparing the three different shape regression approaches of the Point Transformer (Sec. \ref{sec:shape_approaches}), the performance of the direct displacement and heatmap regression are nearly identical ($4\ts\text{mm}$ error and 85\% DSC), with a minor superiority for the direct displacement formulation for smaller structures like the clavicles ($3.8\pm1.7\ts\text{mm}$ vs $3.9\pm1.7\ts\text{mm}$). The direct shape regression performs significantly worse, with a gap of nearly $3.5\ts\text{mm}$ in error and 9\% DSC. We find the Point Transformer outperforms the PointNet in all three approaches (exemplified by the displacement approach with a gap of 5\% DSC).
Taking the ASD distribution (Fig. \ref{fig:asd_distribution}) into consideration, it can be seen that the LR-ASPP has the lowest consistency in its predictions and drops below the performance of the Point Transformer in 10.34\% of all test images.

As an ablation study, we compare the robustness of all models by masking out an area in the input image like in \cite{gaggion_improving_2023}. The position of the mask is chosen randomly, and its size is defined as a fracture of the image size increasing linearly from 0 to 1 with a step size of $0.1$. For each sampling point, we mask the test images and evaluate the models.
Our experiment shows higher robustness of our proposed pipeline to corrupted input than the pixel-level baselines, with the Point Transformer surpassing the PointNet. Considering the decline in relative DSC performance (Fig. \ref{fig:ablation_rel_perf}), the pixel-level baselines drop faster and being outperformed by the shape-based approaches at an image masking of 50\%. However, for the absolute performance (Fig. \ref{fig:ablation_abs_perf}), the Point Transformer performs only best in the case of 60\% corruption. Below this mark, it is outperformed by the nnU-Net and above it, the mean shape scores the highest DSC.

\section{Discussion}
According to the quantitative results in Tab. \ref{tab:quantitave_results}, our proposed pipeline is surpassed by the nnU-Net. However, when using the same lightweight CNN backbone, we can show that its performance is comparable to pixel-level approaches like the LR-ASPP. Note that for shape-based methods, the DSC tends to be underestimated, as some performance is lost due to the conversion from (subpixel) landmark position to mask.  Likewise, for pixel-level baselines, the ASD tends underestimated, as it is calculated on the overall segmentation contour instead of $L$ landmarks.

Our ablation study shows an advantage of our method in terms of higher robustness regarding input corruption. In particular, our shape-based Point Transformer outperforms the LR-ASPP at an early stage of 30\% input corruption (Fig. \ref{fig:ablation_abs_perf}). 
Fig. \ref{fig:port_artefact} demonstrates another superiority of our method, where two ports for injecting medication caused a little domain shift, resulting in a false positive in the LR-ASPP's prediction. Our shape-based method, however, cannot produce such errors due to their anatomical implausibility.

In conclusion, we propose a combination of a CNN backbone and a geometric neural network for shape-based segmentation. We can show that our approach is in reach of the performance of pixel-level methods like the LR-ASPP when the same CNN backbone is used. Furthermore, we demonstrate the benefit of a shape-based approach being less sensitive to image corruption.

In our future work, we will investigate stronger CNN backbones and further optimization of hyperparameters, as well as a possible refinement with a cascade approach \cite{in_young_cascade_2017} to narrow the performance gap to the nnU-Net. Besides that, we plan to take advantage of the geometric understanding of the shape by our model to use it in a human in the loop situation, where it accelerates the annotations process of segmentation mask by allowing to automatically propagate a manual refinement of one single landmark to the whole predicted shape.

\printbibliography

@inproceedings{qi_pointnet_2016,
	author = {Qi, Charles R. and Su, Hao and Mo, Kaichun and Guibas, Leonidas J.},
    title = {PointNet: Deep Learning on Point Sets for 3D Classification and Segmentation},
    booktitle = {Proc IEEE Comput Soc Conf Comput Vis Pattern Recognit},
    month = {July},
    year = {2017}
}

@inproceedings{liang_polytransform_2020,
    author = {Liang, Justin and Homayounfar, Namdar and Ma, Wei-Chiu and Xiong, Yuwen and Hu, Rui and Urtasun, Raquel},
    title = {PolyTransform: Deep Polygon Transformer for Instance Segmentation},
    booktitle = {Proc IEEE Comput Soc Conf Comput Vis Pattern Recognit},
    month = {June},
    year = {2020}
}

@inproceedings{howard_searching_2019,
	author = {Howard, Andrew and Sandler, Mark and Chu, Grace and Chen, Liang-Chieh and Chen, Bo and Tan, Mingxing and Wang, Weijun and Zhu, Yukun and Pang, Ruoming and Vasudevan, Vijay and Le, Quoc V. and Adam, Hartwig},
    title = {Searching for MobileNetV3},
    booktitle = {Proc IEEE Comput Soc Conf Comput Vis Pattern Recognit},
    month = {October},
    year = {2019}
}

@inproceedings{He2016,
	author = {He, Kaiming and Zhang, Xiangyu and Ren, Shaoqing and Sun, Jian},
    title = {Deep Residual Learning for Image Recognition},
    booktitle = {Proc IEEE Comput Soc Conf Comput Vis Pattern Recognit},
    month = {June},
    year = {2016}
}

@article{gaggion_improving_2023,
	  author={Gaggion, Nicolás and Mansilla, Lucas and Mosquera, Candelaria and Milone, Diego H. and Ferrante, Enzo},
    journal={IEEE Trans Med Imaging},
    title={Improving Anatomical Plausibility in Medical Image Segmentation via Hybrid Graph Neural Networks: Applications to Chest X-Ray Analysis},
    year={2023},
    volume={42},
    number={2},
    pages={546-556},
    doi={10.1109/TMI.2022.3224660}
}

@inproceedings{castrejon_annotating_2017,
	author = {Castrejon, Lluis and Kundu, Kaustav and Urtasun, Raquel and Fidler, Sanja},
    title = {Annotating Object Instances With a Polygon-RNN},
    booktitle = {Proc IEEE Comput Soc Conf Comput Vis Pattern Recognit},
    month = {July},
    year = {2017}
}

@inproceedings{zhao_point_2021,
    author    = {Zhao, Hengshuang and Jiang, Li and Jia, Jiaya and Torr, Philip H.S. and Koltun, Vladlen},
    title     = {Point Transformer},
    booktitle = {Proc IEEE Int Conf Comput Vis},
    month     = {October},
    year      = {2021},
    pages     = {16259-16268}
}

@article{shiraishi_development_2000,
	title = {Development of a Digital Image Database for Chest Radiographs With and Without a Lung Nodule: Receiver Operating Characteristic Analysis of Radiologists' Detection of Pulmonary Nodules},
	volume = {174},
	issn = {0361-803X, 1546-3141},
	url = {https://www.ajronline.org/doi/10.2214/ajr.174.1.1740071},
	doi = {10.2214/ajr.174.1.1740071},
	shorttitle = {Development of a Digital Image Database for Chest Radiographs With and Without a Lung Nodule},
	pages = {71--74},
	number = {1},
	journaltitle = {Am J Roentgenol},
	shortjournal = {American Journal of Roentgenology},
	author = {Shiraishi, Junji and Katsuragawa, Shigehiko and Ikezoe, Junpei and Matsumoto, Tsuneo and Kobayashi, Takeshi and Komatsu, Ken-ichi and Matsui, Mitate and Fujita, Hiroshi and Kodera, Yoshie and Doi, Kunio},
	urldate = {2023-11-11},
	date = {2000-01},
	langid = {english},
}

@InProceedings{leibe_human_2016,
	author="Bulat, Adrian
    and Tzimiropoulos, Georgios",
    title="Human Pose Estimation via Convolutional Part Heatmap Regression",
    booktitle="Comput Vis ECCV",
    year="2016",
    publisher="Springer International Publishing",
    address="Cham",
    pages="717--732",
    isbn="978-3-319-46478-7"
}

@article{isensee_nnu-net_2021,
	title = {{nnU}-Net: a self-configuring method for deep learning-based biomedical image segmentation},
	volume = {18},
	issn = {1548-7091, 1548-7105},
	url = {https://www.nature.com/articles/s41592-020-01008-z},
	doi = {10.1038/s41592-020-01008-z},
	shorttitle = {{nnU}-Net},
	pages = {203--211},
	number = {2},
	journaltitle = {Nat Methods},
	author = {Isensee, Fabian and Jaeger, Paul F. and Kohl, Simon A. A. and Petersen, Jens and Maier-Hein, Klaus H.},
	urldate = {2023-11-11},
	date = {2021-02},
	langid = {english},
}

@InProceedings{in_young_cascade_2017,
author="Ha, In Young
and Wilms, Matthias
and Heinrich, Mattias P.",
title="Multi-Object Segmentation in Chest X-Ray Using Cascaded Regression Ferns",
booktitle="Proc BVM",
year="2017",
pages="254--259",
isbn="978-3-662-54345-0"
}

\end{document}